\documentclass[10pt,twocolumn,letterpaper]{article}

\usepackage[pagenumbers]{cvpr} 

\usepackage{graphicx}
\usepackage{amsmath}
\usepackage{amssymb}
\usepackage{booktabs}

\usepackage[pagebackref,breaklinks,colorlinks]{hyperref}

\usepackage{adjustbox}

\usepackage{kotex}
\usepackage{array}
\usepackage{color, colortbl, xcolor}
\usepackage{bm}
\makeatletter

\def\env@cases#1{%
  \let\@ifnextchar\new@ifnextchar
  \left\lbrace\def\arraystretch{1.2}%
  \array{@{}#1@{\quad}l@{}}}
\makeatother
\usepackage{multirow}
\usepackage{pifont}
\definecolor{Gray}{gray}{0.9}
\usepackage{caption}
\usepackage{subcaption}

\usepackage{algorithm}
\usepackage{algorithmic}

\usepackage[capitalize]{cleveref}
\crefname{section}{Sec.}{Secs.}
\Crefname{section}{Section}{Sections}
\Crefname{table}{Table}{Tables}
\crefname{table}{Tab.}{Tabs.}

\def\cvprPaperID{17} 
\def\confName{CVPR}
\def\confYear{2022}

\begin{document}

\title{Pose-MUM : Reinforcing Key Points Relationship \\ for Semi-Supervised Human Pose Estimation}

\author{
JongMok Kim$^{1,2}$~~~
Hwijun Lee$^{1,2}$~~~
Jaeseung Lim$^{1}$~~~
Jongkeun Na$^{1}$~~~
Nojun Kwak$^{2}$~~~
Jin Young Choi$^{2}$~~~
\smallskip
\\
$^1$SNUAILAB~~~
$^2$Seoul National University
}

\maketitle
\begin{abstract}

A well-designed strong-weak augmentation strategy and the stable teacher to generate reliable pseudo labels are essential in the teacher-student framework of semi-supervised learning (SSL). 
Considering these in mind,  to suit the semi-supervised human pose estimation (SSHPE) task, we propose a novel approach referred to as {\it Pose-MUM} that modifies Mix/UnMix (MUM) augmentation~\cite{kim2021mum}.
Like MUM in the dense prediction task, the proposed Pose-MUM makes strong-weak augmentation for pose estimation and leads the network to learn the relationship between each human key points much better than the conventional methods by adding the mixing process in intermediate layers in a stochastic manner.
In addition, we employ the exponential-moving-average-normalization (EMAN)~\cite{cai2021exponential} teacher, which is stable and well-suited to the SSL framework and further boosts the performance.
Extensive experiments on MS-COCO dataset show the superiority of our proposed method by consistently improving the performance over the previous methods following SSHPE benchmark.

\end{abstract}

\section{Introduction}
Recent advances in supervised learning have gathered much evidence for reliability and its diversity has become  paramount. 
However, supervised learning methods face the challenge of requiring a massive amount of data preprocessing.
Specifically, dense prediction tasks such as object detection and semantic segmentation require expensive labeling process and likewise, pose estimation task is also costly. 

To mitigate the difficulties, the studies on semi-supervised learning (SSL) provide productive insights by making use of unlabeled data along with the labeled training data. 
To train a student network using a small amount of labeled data, knowledge distillation methods 
have been developed for effective training of the student network with the help of an experienced teacher network, along with various augmentation methods~\cite{tarvainen2017mean, sohn2020fixmatch, berthelot2019mixmatch, verma2019interpolation, liu2021unbiased, kim2021mum, kim2020structured, french2019semi}, as will be briefly mentioned in Sec. \ref{sec:related_works}. 

Recently, Xie \textit{et al}. \cite{Xie_2021_ICCV} suggested a strong augmentation method for SSHPE (here we call it {\it JC Augment}) in the teacher and student framework. 
The {\it JC Augment} adopts dual model for teacher-student network to decouple the teacher and the student, and Joint Cutout as strong augmentation for simulating hard occlusion of key points by performing Cutout~\cite{ke2018multi, devries2017cutout} at human key points. 
However, as clarified in 
\cite{kim2021mum}, Cutout~\cite{devries2017cutout} augmentation leads to information loss, and the dual network training method used to overcome the instability of EMA requires double training cost.


In this paper, to mitigate the difficulties aforementioned, we propose a new augmentation method to reinforce key points relationship for SSHPE in the teacher-student framework, along with building an improved teacher.
We adopt MUM~\cite{kim2021mum} to the SSHPE task and develop an advanced version, called Pose-MUM. 

MUM augmentation shows a powerful strategy for strong-weak augmentation in dense prediction tasks of SSL~\cite{kim2021mum}. 
However, MUM is performed at the image-level, thus may blind connections between any two key points in the whole backbone forward path. 
To capture the relationship between key points, the Pose-MUM is designed to work additional mixing and unmixing process in the feature-level. 
The augmentation repeats mixing feature tiles in a stochastic fashion through the intermediate layers of the network and makes the key points be associated with one another by giving more chances to be exposed to each other.
Hence the Pose-MUM reinforces the capability of exchanging features of adjacent joints, which is a critical property of the pose estimation task.
In addition, to further improve Pose-MUM, we employ the EMAN~\cite{cai2021exponential} which boosts the efficiency of Pose-MUM by stabilizing the training of the teacher. 

Through ablation study, we verify the augmentation effects of MUM and the superiority of Pose-MUM.
Also we evaluate and verify that Pose-MUM+ (Pose-MUM$+$EMAN) makes synergy of stable teacher and strong data augmentation suitable to pose estimation, resulting in significant performance improvement.

\begin{figure}[t]
\centering
\includegraphics[width=1.03\columnwidth]{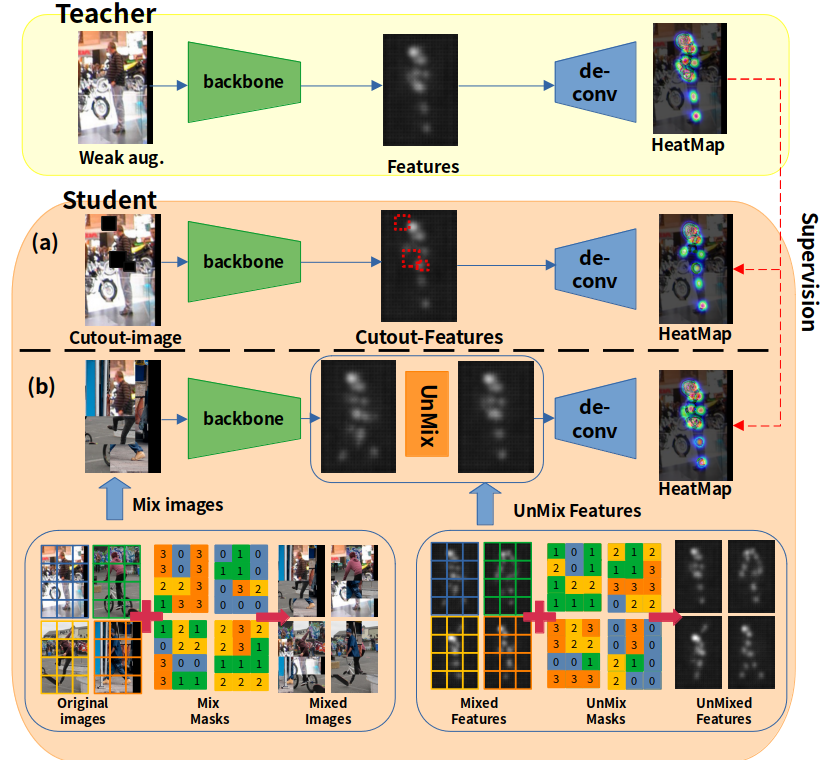}
\caption{System overview of SSHPE based on the teacher-student framework with (a) Joint Cutout and (b) MUM as a strong augmentation.
We set the MUM hyperparmeters, $N_G, N_{T,w}, N_{T,h}$, as 4, 3, and 4, respectively. 
Detailed explanation can be found in Sec.~\ref{sec:method}.
}
\label{fig:JCvsMUM}
\end{figure}

\begin{figure*}[ht]
\centering
\includegraphics[width=0.9\linewidth]{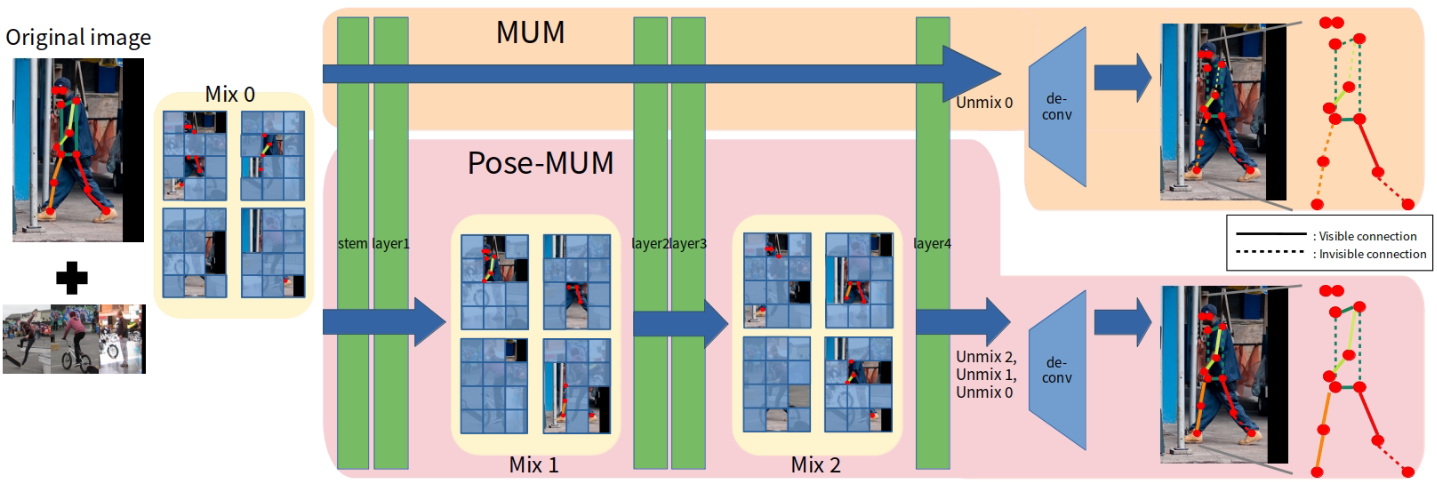}
\caption{{\bf Comparison of MUM and Pose-MUM in the pose estimation task.} Pose-MUM mixes the features in each intermediate layer stochastically. For instance, we perform additional mix operations after layer-1 and layer-3.
Unmixing is performed after layer-4 and reassembles the feature tiles to the original image geometry in the reverse order of mixing.
In the figure, for visibility and simplicity, we insert the image instead of actual feature map and skip the details of mixing and unmixing process. 
In the skeletons of the output image, visible and invisible connections during the training are depicted as solid and dashed lines, respectively.
 }
\label{fig:MUMvsPose-MUM}
\vspace{-0.5cm}
\end{figure*}

\section{Related Works}
\label{sec:related_works}
\noindent{\bf Augmentation methods for SSL in the teacher-student framework.} 
Mean Teacher\cite{tarvainen2017mean} employs the exponential-moving-average (EMA) teacher to generate reliable pseudo-label.  
FixMatch~\cite{sohn2020fixmatch} augments the input image into strong-weak pair to make information gap between the teacher and the student.
MixMatch~\cite{berthelot2019mixmatch} and ICT~\cite{verma2019interpolation} train the student with the interpolated-label to make the student more robust.
Furthermore, Unbiased Teacher~\cite{liu2021unbiased} uses EMA and Focal loss~\cite{lin2017focal} to build a stable teacher.   
MUM~\cite{kim2021mum} is one of the most effective strong-weak data augmentation method in object detection, and the structured consistency loss in \cite{kim2020structured} and the perturbation approach in \cite{french2019semi} also use interpolated-labels for semi-supervised semantic segmentation.
\section{Method}
\label{sec:method}

\subsection{Preliminary}
In the SSHPE task, a set of labeled $\{(x^s_i,y^s_i)\}^{N_s}_{i=1}$ and unlabeled data  $\{x^u_j\}^{N_u}_{j=1}$ is given for training, where $x, y, N$ denote an image, the corresponding label and number of training samples, respectively.
In order to train the student network with the unlabeled data, we generate a pseudo label $\hat{y}^u$ and conduct training as follows:
\begin{equation}\label{eq1}
\mathcal{L}_{total} = \sum_i{\mathcal{L}(x^s_i,y^s_i)} + \lambda_u \cdot \sum_i{\mathcal{L}(x^u_i,\hat{y}^u_i)}, 
\end{equation}
where $\mathcal{L}_{total}$, $\mathcal{L}$ and $\lambda_u$ denote the total loss, loss for human pose estimation task, and weight of the unsupervised learning, respectively.
The overall training process of unsupervised learning is depicted in Figure~\ref{fig:JCvsMUM}.
While {\it JC Augment}~\cite{Xie_2021_ICCV} adopts Joint Cutout, developed using Joint~\cite{ke2018multi} and Cutout~\cite{devries2017cutout}, as a strong augmentation for the SSHPE with teacher-student framework ((a) in Figure~\ref{fig:JCvsMUM}), it hurts the semantic information of key points in the input image and results in too strong augmentation.

\noindent\textbf{MUM.} MUM~\cite{kim2021mum} is a powerful augmentation strategy to leverage unlabeled data. 
It gains an effect of more robust interpolation-regularization with less image information loss in semi-supervised object detection (SSOD).
Furthermore, MUM can be easily extended to SSHPE as shown in Figure~\ref{fig:JCvsMUM}(b).

MUM first requires the number of images to be included per group as $N_G$ and the number of tiles for an image to be split into along the vertical (height) and horizontal (width) direction as $N_{T,h}$ and $N_{T,w}$.
In the mixing phase, each image in a group is split into $N_{T,h}\times N_{T,w}$ tiles and the tiles in the group are mixed (i.g. rearranged) referring the mixing masks which are generated stochastically. 
Mixed features are extracted by the feature extractor. 
In the unmixing phase, unmixed features are created from mixed features according to the unmixing masks which are in the reversed order of mixing masks. 
As shown in Figure \ref{fig:JCvsMUM}(b), natural occlusion of MUM leads to the interpolated-regularization effect, and the adequately reassembled feature map successfully generates the predicted heat map.
Otherwise, {\it JC Augment}~\cite{Xie_2021_ICCV} suffers from the loss of joint information and predicts a bad heat map in Figure \ref{fig:JCvsMUM}(a).

\noindent\textbf{EMA.} A line of recent research builds the teacher~\cite{tarvainen2017mean, kim2021mum, liu2021unbiased} by temporal ensembling of the student's weight via EMA:
\begin{equation}\label{eq2}
\theta^{t+1}_T =  \tau \cdot \theta^t_T + (1-\tau) \cdot \theta^{t+1}_S,
\end{equation}
where $\theta_T$, $\theta_S$, $t$ and $\tau$ denote the parameters of the teacher, the parameters of the student, step, and decay rate, respectively. 
Note that $\theta$ denotes only the trainable parameters, and EMA does not accumulate the statistics of batch normalization (BN) \cite{ioffe2015batch}.
However, recent works~\cite{Xie_2021_ICCV, cai2021exponential} report the unstable aspects of EMA and thus {\it JC Augment}~\cite{Xie_2021_ICCV} fails to stabilize the training process with EMA and bypasses it using a duplicated model.

\subsection{Pose-MUM}
As pose estimation tasks put emphasis on strong attention between key points, intense human occlusion in an image can obfuscate the training, resulting in loss of semantic label information of occluded key points or neighboring one.
Joint Cutout and MUM, which induces the occlusion of connections between the key points in the whole forward path of the backbone, can hinder the training of the student network from learning the relationship between key points.

To tackle this problem, we propose Pose-MUM, which stochastically adds a mixing process in each intermediate layer of the backbone and the corresponding unmixing process in a predefined unmixing location as described in Figure~\ref{fig:MUMvsPose-MUM}.
Since Pose-MUM increases the possibility of reuniting the adjacent tiles and joints, it relaxes the hard occlusion of previous methods.
For instance, MUM loses the chance to capture the relationship between left shoulder and elbow, but Pose-MUM captures the relationship by stochastic mixing as shown in Figure~\ref{fig:MUMvsPose-MUM}.
At the same time, Pose-MUM also induces weak occlusions for most connections of key points, generating robust but more easily learnable
augmented images.
The overall operation of Pose-MUM is summarized as follows.
\begin{itemize}
    \item A group of $N_G$ input images split into $N_{T,h}\times N_{T,w}$ tiles (sub-images) are mixed in the same way as MUM operation.
    \item The additional feature-level mixing processes at the intermediate layers of the encoder are done stochastically in a uniform distribution. 
    The corresponding mixing masks are stacked. 
    Note that Pose-MUM consists of additional feature-level mixing stages for more robust key points bond, while MUM does not.
    \item For unmixing the outputs of the encoder with the stacked masks, we restore the image geometry from the mixed feature maps by using the stacked masks in the reverse order at the end of the encoder.
\end{itemize}

\subsection{EMAN}
\label{sec:eman}

A wide range of fields uses EMA~\cite{tarvainen2017mean} to improve the performance by co-training the teacher and the student.
EMA plays an important role of building a more natural teacher network especially in SSL which has to leverage unlabeled data.
However, in the recent augmentation study~\cite{Xie_2021_ICCV} in SSHPE, EMA is shown to perform worse than the \textit{Single} model in which the teacher and the student share the same trainable weight, and the training fails to converge.
To overcome the failure, a dual network architecture has been used to decouple the teacher and the student. 
The dual architecture makes the cost double, thus leads to  inefficient learning.

To tackle this problem, we employ EMAN \cite{cai2021exponential} to train a stable and superior teacher for the SSHPE.
Unlike the standard BN in EMA, where the statistics are computed within each batch, EMAN updates teacher's statistics by exponential moving average from the student's statistics.
Therefore, EMAN updates not only trainable parameters in Eq.  (\ref{eq2}), but also the BN statistics as follows:
\begin{equation} \label{eq3}
\begin{split}
\mu^{t+1}_T =  \tau \cdot \mu^t_T + (1-\tau) \cdot \mu^{t+1}_S ,\\
\sigma^{t+1}_T =  \tau \cdot \sigma^t_T + (1-\tau) \cdot \sigma^{t+1}_S ,\\
\end{split}
\end{equation}
where $\mu_T$, $\mu_S$, denote the running means in BN of the teacher and the student, whereas $\sigma_T$ and $\sigma_S$ are the running standard deviations in BN of the teacher and the student, respectively. 

\begin{figure}[t] 
\centering
\begin{minipage}{.5\linewidth}
    \centering
    \includegraphics[width=1.\linewidth, height = 0.75\linewidth]{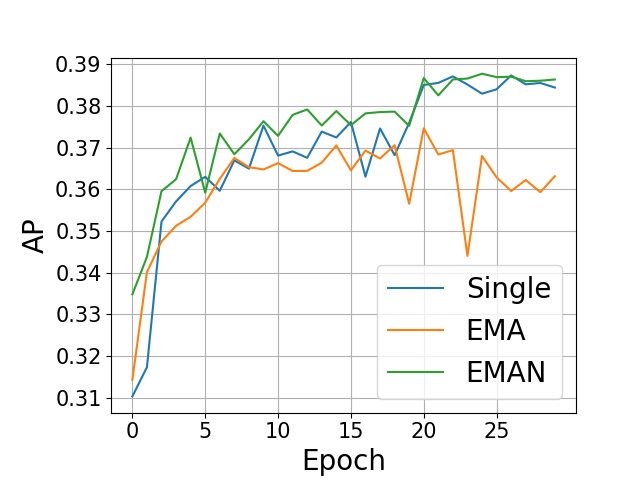}
    \caption{AP performance}
    \label{fig:eman_ap}
\end{minipage}%
\begin{minipage}{.5\linewidth}
    \centering
    \includegraphics[width=1.\linewidth,  height = 0.75\linewidth]{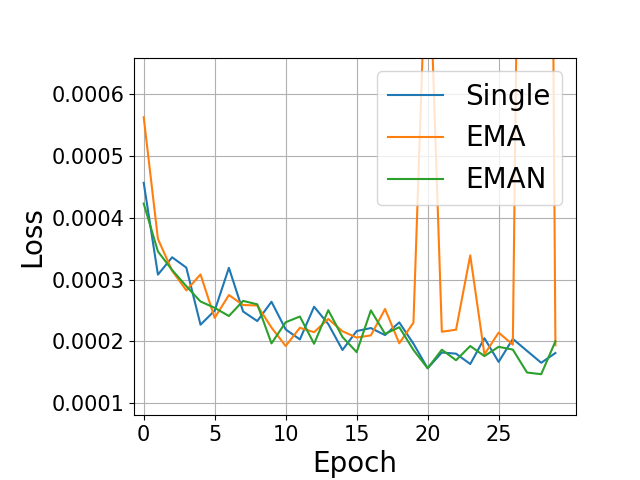}
    \caption{Training loss}
    \label{fig:eman_loss}
\end{minipage}
\vspace{-0.1cm}
\end{figure}

Figures \ref{fig:eman_ap} and \ref{fig:eman_loss} demonstrate the efficacy of EMAN compared with Single and EMA with affine augmentation~\cite{Xie_2021_ICCV}.
As shown in the figures, EMAN shows the most stable training curve. 
Note that the instability of EMA comes from the cross-sample dependency and the model parameter mismatch  \cite{cai2021exponential}.
In the figures, although the training losses of EMAN and EMA are quite similar before epoch 20, EMAN shows better performance than EMA.
The reason originates from the claim that batch normalization in EMA causes information leakage, which provides noisy label information while training the student \cite{cai2021exponential}.
In addition, the training curves of EMA become very unstable after learning rate decay (at epoch 20 and 25). The reason is due to the enlarged BN parameter mismatch.
On the other hand, the training curves for EMAN become stable because EMAN overcomes the two issues of EMA by averaging the batch statistics of the student. 

 \section{Experiments}
 \subsection{Experimental Settings}
\noindent\textbf{Datasets.}
We used MS-COCO dataset \cite{lin2014microsoft} to evaluate our method. The dataset is divided into parts such that each part consists of random 1K, 5K, and 10K labels follows the previous work~\cite{Xie_2021_ICCV}. 

\noindent\textbf{Evaluation.}
We recorded the mean average precision (AP) with 10 object key points similarity (OKS) thresholds.
All of our evaluations follow the recent augmentation method \cite{Xie_2021_ICCV} to verify the effectiveness of our proposed method in a fair comparison. 


\noindent\textbf{Implementations.}
SimpleBaseline \cite{xiao2018simple} and ResNet18 \cite{he2016deep} are used as our default HPE model and the backbone, respectively.  
The input image size is set to 192$\times$256. We also follow the training details of \cite{Xie_2021_ICCV}:
ADAM as optimizer, initial learning rate of $1e^{-3}$, and learning rate decay to $1e^{-4}$ and $1e^{-5}$ at 20 and 25 epochs out of 30 for 1K, and 70 and 90 epochs out of 100 epochs for 5K/10K.
For ablation study and main results, we use ResNet-18 \& 1K label and ResNet-18, -101 \& 1K, 5K, 10K label setup respectively as default.

\begin{table}[t]
\centering
\caption{Analysis on hyper-parameters of MUM for SSHPE task.
}
\begin{tabular}{|l|c|l|c|}
\hline
Augmentation                                                              & \multicolumn{1}{l|}{$N_G$} & Unmixing location & \multicolumn{1}{l|}{AP (\%)} \\ \hline
Affine $+$ MUM                                                                         & 4                       & After Encoder           & \textbf{45.5}                         \\ 
Affine $+$ MUM                                                                         & 2                       & After Encoder           & 44.6                         \\ \hline
Affine $+$ MUM                                                                         & 4                       & After Layer-2            & 39.2                         \\
Affine $+$ MUM                                                                         & 4                       & After Decoder           & 45                           \\ \hline
\end{tabular}
\label{tab:mum_hyper_para}
\vspace{-0.1cm}
\end{table}
 

\subsection{Hyper-Parameter Analysis}

\noindent\textbf{Hyper-parameters of MUM for SSHPE task.} MUM augmentation was reported as a powerful strategy for strong-weak augmentation in dense prediction tasks of SSL~\cite{kim2021mum}. 
At the first time, we tried to apply MUM to the SSHPE task and developed an advanced version, called Pose-MUM, for SSHPE task. 
To this end, we conducted an experiment to determine hyper-parameters of MUM for the affine~\cite{thewlis2017unsupervised} augmentation based method~\cite{Xie_2021_ICCV}: the number of images per group for mixing $N_G$ and the unmixing location where reassembling is performed. 
As shown in Table~\ref{tab:mum_hyper_para}, the best AP result is obtained when $N_G=4$ and the ummixing process is located after the encoder. 
In the remaining experiments, we use this configuration for both MUM and Pose-MUM.

\noindent\textbf{Decay Rates for Training.} 
In the subsection \ref{sec:eman}, we discussed the superiority of EMAN to EMA. 
To confirm the claim, we compared the AP performance of Single, EMA, and EMAN with the basic affine-only model. 
The results are given in Table \ref{tab:ema}.
As mentioned earlier, the EMA model using a commonly-used decay rate (0.999) did not show satisfactory performance as shown in Table \ref{tab:ema}.
When the decay rate decreases to 0.6, the performance improved to a certain extent, but it could not reach that of the baseline model (\textit{Single}) due to the drawbacks of EMA.
When EMAN was used with the same decay rate, it produced a slight performance improvement over that of the \textit{Single} baseline and the training became stable.
As discussed in \cite{Xie_2021_ICCV} because of the nature of the teacher-student framework, it is more beneficial for a teacher to be decoupled from the student, thus we adopted EMAN as our proposed method.

\begin{table}[h!]
\caption{Performance of Affine baseline \cite{Xie_2021_ICCV} with Single, EMA, and EMAN. 
Note that Single denotes the weight of the teacher is equivalent to that of the student.}
\centering
\begin{tabular}{|l|l|c|}
\hline
Model           & decay rate & \multicolumn{1}{l|}{AP (\%)} \\ \hline
\cite{Xie_2021_ICCV} $+$ Single & -          & 38.44                        \\
\cite{Xie_2021_ICCV} $+$  EMA     & 0.999      & 35.94                        \\
\cite{Xie_2021_ICCV} $+$ EMA    & 0.6        & 36.31                        \\
\cite{Xie_2021_ICCV} $+$ EMAN    & 0.6        & 38.63                        \\ \hline
\end{tabular}
\label{tab:ema}
\vspace{-0.1cm}
\end{table}

\begin{table*}[t!]
\caption{Ablation results of MUM and Pose-MUM for SSHPE task in the single model teacher-student framework. $+$ means the augmentation component is used in SSHPE experiment.
Note that -s denotes single model framework.}
\centering
\small
\begin{tabular}{|l|c|c|c|c|c|c|c|}
\hline
Variants & Affine Transform \cite{thewlis2017unsupervised}& Rand Augment \cite{cubuk2020randaugment}& Joint~\cite{ke2018multi} Cutout~\cite{devries2017cutout}& MUM~\cite{kim2021mum} & Pose-MUM & {AP(\%)} \\ \hline
Supervised Baseline \cite{xiao2018simple} & & && & & 31.5 \\
Affine Augment-s \cite{Xie_2021_ICCV}  
&$+$ & & & & & 38.5   \\
JC Augment-s$^{\dagger}$ \cite{Xie_2021_ICCV} 
&$+$& & $+$ & & &  42.1 \\ \hline
MUM-JC   & $+$ & &$+$ & $+$ & & 41.9 \\ 
MUM-RA  &$+$ &$+$ && $+$ &  & 43.8 \\ 
MUM  &$+$ & & &$+$ && 45.5 \\ \hline
Pose-MUM-AT   & $+$ & & & & $+$&  {\textbf{46.12}} \\
Pose-MUM & & &  & &$+$ & {\textbf{46.48}} \\ \hline
\end{tabular}
\vspace{0.3cm}
$\dagger$: {\it JC Augment-s} shows the {\it best} accuracy among augmentation methods for SSHPE in the single model baseline framework \cite{xiao2018simple}.
\label{tab:pose_mum}
\vspace{-0.5cm}
\end{table*}


\subsection{Ablation study}

For the ablation study, we prepared variant models (see Table~\ref{tab:pose_mum}) by a combination of existing augmentation methods, MUM, and Pose-MUM introduced in this work.
In \cite{Xie_2021_ICCV}, Affine \cite{thewlis2017unsupervised} and Joint Cutout ~\cite{ke2018multi, devries2017cutout} augmentation techniques were added to the simple baseline model \cite{xiao2018simple}, in which AP increased to 42.1\%.
When we added MUM to this model, the AP decreased by 0.2\%p. 
However, when we used only MUM without both Joint Cutout and Rand Augment, we achieved an improved AP of 45.5\%. 
By replacing MUM with Pose-MUM, AP was additionally improved by 0.62\%p. 
Interestingly, when using only Pose-MUM without affine augmentation, we achieved the best accuracy of 46.48\%, which was an approximately 14.98\%p improvement from that of the Supervised Baseline \cite{xiao2018simple}, and a 4.38\%p improvement from the best AP of the same structure in \cite{Xie_2021_ICCV}. 

To the best of our knowledge, this is the largest improvement obtained only by using data augmentation. 
This result implies that Pose-MUM doing feature-level stochastic mixing/unmixing can provide beneficial augmentation effects and greatly contribute to the performance improvement.

\subsection{Synergy of Pose-MUM and EMAN}

In this subsection, the synergy of Pose-MUM and  EMAN is evaluated and Pose-MUM$+$ (Pose-MUM$+$EMAN) is verified to be effective for ResNet-101 as well as ResNet-18 and for various size of labeled datasets (1K, 5K, 10K). 
First, we investigated the decay rate of EMAN depending on the backbone networks adopted for Pose-MUM$+$.
Unlike the previous SSL studies based on the teacher-student framework~\cite{liu2021unbiased, tarvainen2017mean}, both models experienced performance degradation at large decay rates, 
as shown in Table \ref{tab:eman}.
The best decay rate is different depending on the backbone networks, where the best one is 0.6 for ResNet-18 and 0.9 for ResNet-101.
In particular, ResNet-18 achieves 46.9\% AP which is an improvement by 0.42\%p from that of Pose-MUM which is 46.48\%. Note that this the state-of-the-art result, advancing the AP around 4.8\%p upward from the current state-of-the-art.
By upgrading the backbone to ResNet-101, Pose-MUM$+$ achieves high accuracy of 48.94\%.

\begin{table}[h!]
\centering
\caption{AP performance of Pose-MUM$+$ with varying EMAN decay rate for ResNet-18 and ResNet-101 backbones.}
\begin{tabular}{|l|c|c|}
\hline
EMAN decay rate & \multicolumn{1}{l|}{ResNet-18} & \multicolumn{1}{l|}{ResNet-101} \\ \hline
0.999           & 44.33                          & 46.13                           \\
0.99            & 45.13                          & 47.29                           \\
0.9             & 46.47                          & \textbf{48.94}                  \\
0.6             & \textbf{46.9}                  & 47.64                           \\
0.5             & 45.91                          & 48.8                            \\ \hline
\end{tabular}
\label{tab:eman}
\end{table}

Lastly, the proposed Pose-MUM$+$ was tested in two backbones and three data split strategies.
There were performance improvements from a minimum of 2.1\%p to a maximum of 4.8\%p over the protocols.
Since SSL technique shows a notable impact due to the lack of label data, the performance tends to be easily improved on 1K dataset. 
In our results, it is quite impressive that the same performance improvement can be achieved by Pose-MUM+ on 5K and 10K datasets although SSL techniques are less potent on large labeled datasets.
The generality of the proposed Pose-MUM+ is confirmed by verifying the performance improvement in various models and environments.
In consequence, the synergy of stable teacher and strong data aggregation suitable for pose estimation can lead to a significant performance gain.  

\begin{table}[t!]
\centering
\caption{AP improvement of the proposed Pose-MUM$+$ from the the JC Augment-s on different data protocols of semi-supervised human pose estimation. We tested them in three data splits and two  backbones, ResNet-18 and ResNet-101.}
\begin{tabular}{|l|l|ccc|}
\hline
\multirow{2}{*}{Method}   & \multirow{2}{*}{Backbone} & \multicolumn{3}{l|}{Amount of labeled data}                                \\
                          &                           & \multicolumn{1}{l}{1K} & \multicolumn{1}{l}{5K} & \multicolumn{1}{l|}{10K} \\ \hline
JC   & ResNet-18                 & 42.1                   & 52.3                   & 57.3                     \\
Augment-s \cite{Xie_2021_ICCV}  & ResNet-101                & 46.86                  & 56.7                   & 62.32                    \\ \hline
\multirow{2}{*}{Pose-MUM$+$}     & ResNet-18                 & 46.9                   & 56.46                  & 59.53                    \\
                 & ResNet-101                & 48.94                  & 60.72                  & 64.65                    \\ \hline
\end{tabular}
\label{tab:main}
\vspace{-0.3cm}
\end{table}


\section{Conclusion}
This paper proposed a comprehensive approach of generating a better strong-weak data pair for human pose estimation in the SSL framework, called Pose-MUM. 
The proposed Pose-MUM has additional mixing/unmixing operations in the intermediate layers and makes the student network catch more connections between human key points.
Therefore, Pose-MUM successfully generates a moderately strong augmented input image from the perspective of the connection of adjacent key points, which is essential for the success of the human pose estimation task.
By adopting EMAN for the stable teacher and meticulously tuning the decay rate, we could improve the accuracy by a large margin. 
In conclusion, the proposed Pose-MUM+ outperforms the current SOTA model having the same SSHPE framework with ours on the MS-COCO dataset.

{\small
\bibliographystyle{ieee_fullname}
\bibliography{egbib}
}

\end{document}